\documentclass{article}

\usepackage{microtype}
\usepackage{graphicx}
\usepackage{booktabs} % for professional tables

\usepackage{hyperref}

\usepackage[preprint]{icml2026}

\usepackage{amsmath}
\usepackage{amssymb}
\usepackage{mathtools}
\usepackage{amsthm}

\usepackage[capitalize,noabbrev]{cleveref}

\theoremstyle{plain}

\theoremstyle{definition}

\theoremstyle{remark}

\usepackage[textsize=tiny]{todonotes}

\usepackage{enumitem}
\usepackage{graphicx}
\usepackage{algorithm,algorithmic}

\usepackage{arydshln}
\usepackage{subfigure}
\usepackage{tcolorbox}  % 文本加框
\tcbuselibrary{breakable}   % 配合文本加框，使得框可以跨页
\usepackage{pifont}
\usepackage{caption}
\usepackage{multirow}
\usepackage{hyperref}       % hyperlinks
\usepackage{balance}

\usepackage{color, xspace}

\usepackage[bottom]{footmisc}
\usepackage{tcolorbox}  % 文本加框
\usepackage[dvipsnames]{xcolor}

\newcommand{\name}{IMAGINE\xspace}

\icmltitlerunning{Navigate the Unknown: Enhancing LLM Reasoning with Intrinsic Motivation Guided Exploration}

\begin{document}

\twocolumn[
  \icmltitle{Navigate the Unknown: Enhancing LLM Reasoning with \\ Intrinsic Motivation Guided Exploration}

  % It is OKAY to include author information, even for blind submissions: the
  % style file will automatically remove it for you unless you've provided
  % the [accepted] option to the icml2026 package.

  % List of affiliations: The first argument should be a (short) identifier you
  % will use later to specify author affiliations Academic affiliations
  % should list Department, University, City, Region, Country Industry
  % affiliations should list Company, City, Region, Country

  % You can specify symbols, otherwise they are numbered in order. Ideally, you
  % should not use this facility. Affiliations will be numbered in order of
  % appearance and this is the preferred way.
  \icmlsetsymbol{clar}{*}
  \icmlsetsymbol{corr}{†}

  \begin{icmlauthorlist}
    \icmlauthor{Jingtong Gao}{cityu,clar}
    \icmlauthor{Ling Pan}{hkust}
    \icmlauthor{Yejing Wang}{cityu}
    \icmlauthor{Rui Zhong}{kuai}
    \icmlauthor{Chi Lu}{kuai}
    \icmlauthor{Maolin Wang}{cityu}
    \icmlauthor{Qingpeng Cai}{kuai,corr}
    \icmlauthor{Peng Jiang}{kuai}
    \icmlauthor{Xiangyu Zhao}{cityu,corr}
  \end{icmlauthorlist}

  \icmlaffiliation{cityu}{City University of Hong Kong}
  \icmlaffiliation{hkust}{Hong Kong University of Science and Technology}
  \icmlaffiliation{kuai}{Kuaishou Technology}

  \icmlcorrespondingauthor{Qingpeng Cai}{caiqingpeng@kuaishou.com}
  \icmlcorrespondingauthor{Xiangyu Zhao}{xianzhao@cityu.edu.hk}

  % You may provide any keywords that you find helpful for describing your
  % paper; these are used to populate the "keywords" metadata in the PDF but
  % will not be shown in the document
  \icmlkeywords{Machine Learning, ICML}

  \vskip 0.3in
]

\printAffiliationsAndNotice{*This work was completed at Kuaishou Technology. †Corresponding authors.}

% this must go after the closing bracket ] following \twocolumn[ ...

% This command actually creates the footnote in the first column listing the
% affiliations and the copyright notice. The command takes one argument, which
% is text to display at the start of the footnote. The \icmlEqualContribution
% command is standard text for equal contribution. Remove it (just {}) if you
% do not need this facility.

% Use ONE of the following lines. DO NOT remove the command.
% If you have no special notice, KEEP empty braces:
% \printAffiliationsAndNotice{}  % no special notice (required even if empty)
% Or, if applicable, use the standard equal contribution text:
% \printAffiliationsAndNotice{\icmlEqualContribution}

\begin{abstract}
  Reinforcement Learning (RL) has become a key approach for enhancing the reasoning capabilities of large language models.
  However, prevalent RL approaches like proximal policy optimization and group relative policy optimization suffer from sparse, outcome-based rewards and weak exploration incentives, limiting their effectiveness.
  Specifically, sparse rewards offer limited feedback, especially on difficult problems, and introduce biases favoring familiar trajectories over novel reasoning paths.
  These issues critically undermine performance on complex tasks that inherently require iterative reasoning.
  To overcome these challenges, we propose \textbf{I}ntrinsic \textbf{M}otiv\textbf{A}tion \textbf{G}uided explorat\textbf{I}o\textbf{N} for \textbf{E}nhanced reasoning (\name), which delivers dense rewards and encourages exploration.
  \name introduces three innovations: a trajectory-aware exploration reward that reduces token-level bias efficiently; an error-conditioned reward allocation that promotes efficient exploration on hard samples while stabilizing training; and an advantage-preserving integration mechanism that retains distributional integrity during learning.
  Experiments on four public datasets show that \name improves performance by 22.23\% on AIME 2024. The source code is available to support reproducibility~\footnote{https://github.com/Gaojingtong/IMAGINE}.
\end{abstract}

\begin{sloppypar} 
\section{Introduction}
Reinforcement Learning (RL)~\citep{arulkumaran2017deep,shakya2023reinforcement,ladosz2022exploration} has become essential for training Large Language Models (LLMs)~\citep{hadi2023survey,min2023recent}, evolving from Proximal Policy Optimization (PPO)~\citep{schulman2017proximal} to Group Relative Policy Optimization (GRPO)~\citep{shao2024deepseekmath}. Frameworks like DeepSeek~\citep{guo2025deepseek,liu2024deepseek,shao2024deepseekmath} showcase State-Of-The-Art (SOTA) methods that optimize multi-step reasoning using trajectory-level advantage estimation. RL uniquely transforms sparse rewards~\citep{gao2024designing} into learning gradients, enabling coherent Chain-of-Thought (CoT) reasoning~\citep{wei2022chain,plaat2024reasoning}. By refining action distributions through feedback, RL bridges the gap between linguistic competence and goal-directed problem-solving in LLMs.

While contemporary RL methods like PPO and GRPO demonstrate progress through outcome-based rewards~\citep{uesato2022solving,lightman2023let,wang2024arithmetic}, two fundamental limitations persist for reasoning tasks. First, the reliance on static reward functions creates a sparse learning signal~\citep{bayat2025steering,cao2024beyond} that fails to guide intermediate reasoning steps due to the limited meaningful rewards on hard samples and early training stages, forcing models to navigate vast action spaces with terminal outcome feedback alone{\color{red}~\citep{pmlr-v260-bougie25a,song2025outcomeExploration,bamba2025xrpo}}. This is particularly evident on difficult samples and datasets~\citep{liu2024knowing} where outcome-based rewards are more likely to be zero. Second, despite GRPO's group-wise sampling strategy, the inherent reward structure disincentivizes genuine exploration -- trajectories yielding identical final outcomes receive equivalent advantage estimates regardless of different reasoning CoTs, effectively penalizing computationally intensive sampling efforts~\citep{yu2025dapo,liu2025understanding}. This creates a paradoxical scenario where models optimize for reward exploitation at the expense of systematic exploration, particularly detrimental in difficult reasoning tasks that require complicated reasoning processes.

Traditional exploration methods (e.g., RND~\citep{burda2018exploration}, ICM~\citep{pathak2017curiosity}, Count-Based Exploration~\citep{ostrovski2017count}) promote novelty-seeking via intrinsic rewards, motivated by cognitive theories of curiosity. However, once exploration is moved from embodied control to LLM reasoning, the usual assumptions begin to break: ``progress'' is judged on complete responses and shrinks with problem difficulty, and the state space is combinatorial. In this setting, directly reusing token-level exploration can unintentionally introduce bias towards lengthy responses due to unbalanced sampling, inflate per-token computation when reasoning becomes long, and spread exploration over superficial next tokens rather than sequence-level semantics~\citep{zhou2024balancing,wen2024reinforcing}.

A second challenge arises when integrating intrinsic rewards with advanced Reinforcement Learning with Verifiable Rewards (RLVR) objectives. Methods such as GRPO~\citep{shao2024deepseekmath} rely on carefully normalized advantage estimates to ensure stable policy updates. When exploration rewards are mixed into the same update stream too early, they can alter subsequent advantage statistics in undesirable ways: sometimes punish trajectories simply because less ``novelty'' may reverse the update direction in advantage estimation. Subsequently, instead of serving as an incentive guide through sparse reward spaces, such exploration can become a competing and distorting objective that alters optimization dynamics.

In this paper, we propose an \textbf{I}ntrinsic \textbf{M}otiv\textbf{A}tion \textbf{G}uided explorat\textbf{I}o\textbf{N} for \textbf{E}nhanced reasoning (\name), which makes intrinsic motivation compatible with the way reasoning is expressed and optimized in RL-trained LLMs. \name replaces token-centric novelty with sequence-level signals by using two lightweight trajectory-aware networks to characterize the uniqueness of complete reasoning paths, avoiding both length bias and token-wise overhead. It further focuses exploration where it is most useful by allocating intrinsic incentives only to incorrect trajectories, creating a self-regulating pressure that concentrates on hard cases while remaining stable as the model improves. Finally, \name composes intrinsic incentives with outcome learning in a gradient-safe manner: exploration bonuses are introduced only after advantage computation, preserving the outcome-driven advantage distribution while still steering the search toward diverse reasoning strategies. Together, these choices produce dense, stable, and computationally efficient exploration rewards that integrate cleanly into RL-based LLM training pipelines.

Our contributions are as follows. We introduce \name, an exploration method for RL-based LLM reasoning that provides dense intrinsic guidance without sacrificing training stability or computational efficiency. The method couples trajectory-aware exploration modeling with an error-conditioned reward allocation scheme and an advantage-preserving integration strategy, yielding exploration signals that stay aligned with outcome optimization. Empirically, across four public datasets and two base LLMs (Qwen2.5-3B~\citep{qwen2.5} and DeepSeek-7B~\citep{deepseekai2025deepseekr1incentivizingreasoningcapability}), \name consistently improves LLM's reasoning performance under multiple RL algorithms, including PPO~\citep{schulman2017proximal}, GRPO~\citep{shao2024deepseekmath}, and DAPO~\citep{yu2025dapo}. Notably, the gains are strongest on more difficult benchmarks, suggesting that such sequence-level intrinsic rewards are particularly effective at helping pretrained policies overcome the learning barriers that arise in challenging reasoning instances under RLVR fine-tuning.

\section{Method}\label{sec:method}

Here we first introduce the commonly used RL methods for LLM, i.e., PPO and GRPO, then introduce our method \name.

\subsection{Preliminary}

\begin{figure*}[t]
    \centering
    \includegraphics[width=0.7\linewidth]{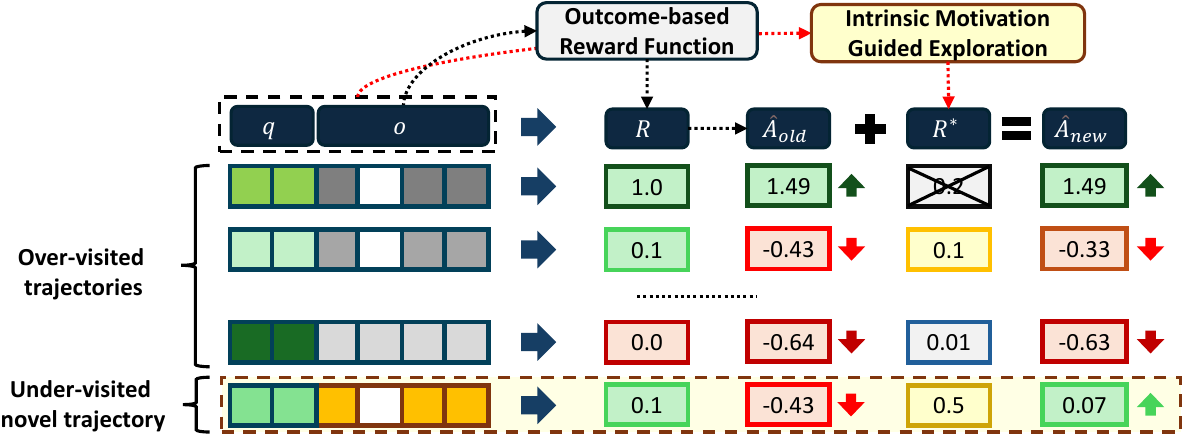}
    \vspace{-1mm}
    \caption{\name. Values with green/yellow boxes denote higher outcome rewards/advantages and exploration rewards; red/blue denote lower values. Black boxes mark the excluded values. $\hat{A}_{old}$ and $\hat{A}_{new}$ denote advantages derived from outcome rewards and \name.}
    \vspace{-4mm}
    \label{fig:motivation}
\end{figure*}

\subsubsection{Proximal Policy Optimization (PPO)}
PPO enhances policy optimization through a clipped objective function that ensures stable updates:
\begin{equation}
\begin{split}
\mathcal{J}_{\mathrm{PPO}}(\theta)=\mathbb{E}_{(q, a) \sim \mathcal{D}, o_{\leq t} \sim \pi_{\theta_{\text {old }}}(\cdot \mid q)}[
\min (w_{t}(\theta) \hat{A}_t, \\
\operatorname{clip}(w_{t}(\theta), 1-\varepsilon, 1+\varepsilon) \hat{A}_t)]
\end{split}
\end{equation}
where
$w_{t}(\theta)=\frac{\pi_\theta(o_{t} \mid q, o_{<t})}{\pi_{\theta_{\text {old }}}(o_{t} \mid q, o_{<t})}$.
Here, $(q,a)$ denotes question-answer pairs from dataset $\mathcal{D}$. $o_t$ and $o_{<t}$ represent the response token at position $t$ and all response tokens before $t$. The current and previous policies are parameterized by $\pi_\theta$ and $\pi_{\theta_{\text{old}}}$ respectively. The advantage estimator $\hat{A}_t$ employs Generalized Advantage Estimation (GAE)~\citep{schulman2015high} with an outcome-based reward function $R$ and a value function $V$, while $\varepsilon$ controls the clipping range. By treating question-response prefixes $[q,o_{<t}]$ as distinct states and constraining policy updates within a trust region, PPO stabilizes RL training for LLMs. However, the joint optimization of policy and value functions introduces computational overhead, limiting training efficiency.

\subsubsection{Group Relative Policy Optimization (GRPO)}
GRPO extends PPO through group-wise advantage normalization, formulated as:
\begin{equation}
\begin{split}
\mathcal{J}_{\mathrm{GRPO}}(\theta)  =\mathbb{E}_{(q, a) \sim \mathcal{D},\{o_i\}_{i=1}^G \sim \pi_{\theta_{\text {old }}(\cdot \mid q)}} 
 [\frac{1}{G} \sum_{i=1}^G \frac{1}{|o_i|} \sum_{t=1}^{|o_i|}(\\
 \min (w_{i, t}(\theta) \hat{A}_{i}, \operatorname{clip}(w_{i, t}(\theta), 1-\varepsilon, 1+\varepsilon) \hat{A}_{i})\\
 -\beta D_{\mathrm{KL}}(\pi_\theta \| \pi_{\mathrm{ref}}))].
\end{split}
\end{equation}
where $w_{i, t}(\theta)=\frac{\pi_\theta(o_{i, t} \mid q, o_{i,<t})}{\pi_{\theta_{\text {old }}}(o_{i, t} \mid q, o_{i,<t})}$ and
$\hat{A}_{i}=\frac{R_i-\operatorname{mean}(\{R_i\}_{i=1}^G)}{\operatorname{std}(\{R_i\}_{i=1}^G)}$.
$G$ responses $\{o_i\}_{i=1}^G$ are generated per input, $\pi_{\mathrm{ref}}$ denotes the reference policy, and $\beta$ controls KL penalty strength. GRPO replaces value function estimation with group-wise normalization on outcome-based rewards, improving training efficiency and effects through reduced computational complexity and group-wise sampling process. However, in most reasoning settings, the outcome reward is accuracy-based~\cite{guo2025deepseek}; for hard samples, most rollouts are incorrect and receive $0$~\cite{yu2025dapo}, which provides limited training guidance within a group. As training progresses, these methods will quickly converge to similar response sequences and no longer provide novel answers, which may also lead to suboptimal performance~\cite{xie2025unlocking}.

\begin{figure}[t]
% \vspace{-1mm}
    \centering
    \includegraphics[width=\linewidth]{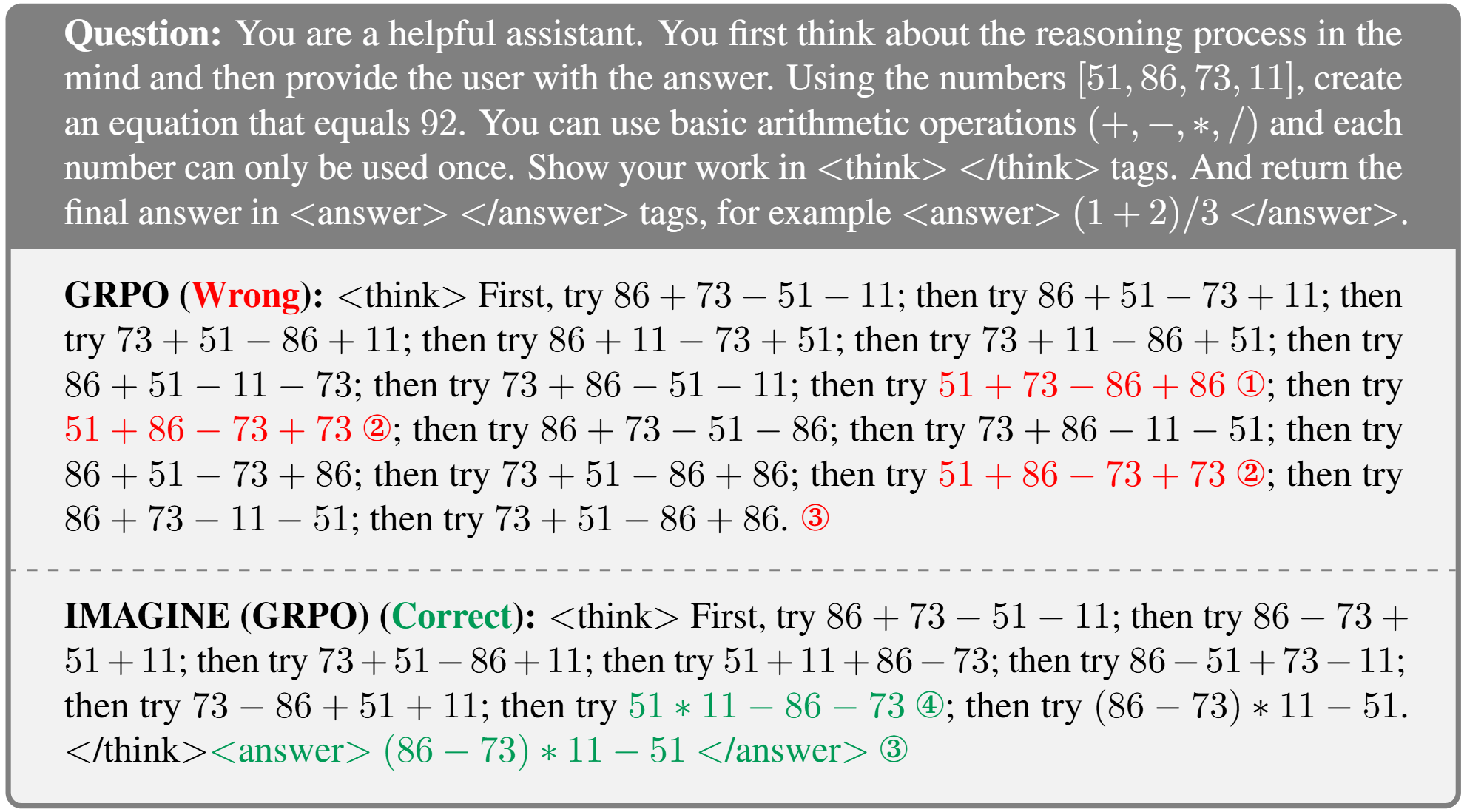}
    \vspace{-3mm}
    \caption{Case study on Countdown-4~\citep{gandhi2024stream} with \underline{simplified responses}. Numbers with circles (e.g., \ding{172}) highlight key differences. Detailed case study is provided in Appendix~\ref{sec:detailcase}.}
    \label{fig:case}
    \vspace{-4mm}
\end{figure}

\subsection{\name}\label{sec:methoddetail}
As previously stated, despite advancements, PPO and GRPO face two key limitations: (1) Sparse reward signals that provide limited training guidance, and (2) weak exploration mechanisms during policy updates. These issues hinder learning efficiency and final optimization. To address these challenges, we propose %Guided Exploration for Reasoning Optimization
an
\textbf{I}ntrinsic \textbf{M}otiv\textbf{A}tion \textbf{G}uided explorat\textbf{I}o\textbf{N} for \textbf{E}nhanced reasoning (\name) method, which introduces structured exploration to enhance LLM reasoning and can be easily implemented in SOTA RL methods like PPO and GRPO. 
As shown in Figure~\ref{fig:motivation}, vanilla RL methods with outcome-based rewards focus only on correct trajectories, ignoring under-visited ones and causing LLMs to converge suboptimally with sparse rewards. \name additionally rewards incorrect under-visited trajectories, preventing the model from getting stuck via intrinsic motivation guided exploration. 

\begin{figure}[t]
% \vspace{-1mm}
    \centering
    \includegraphics[width=0.8\linewidth]{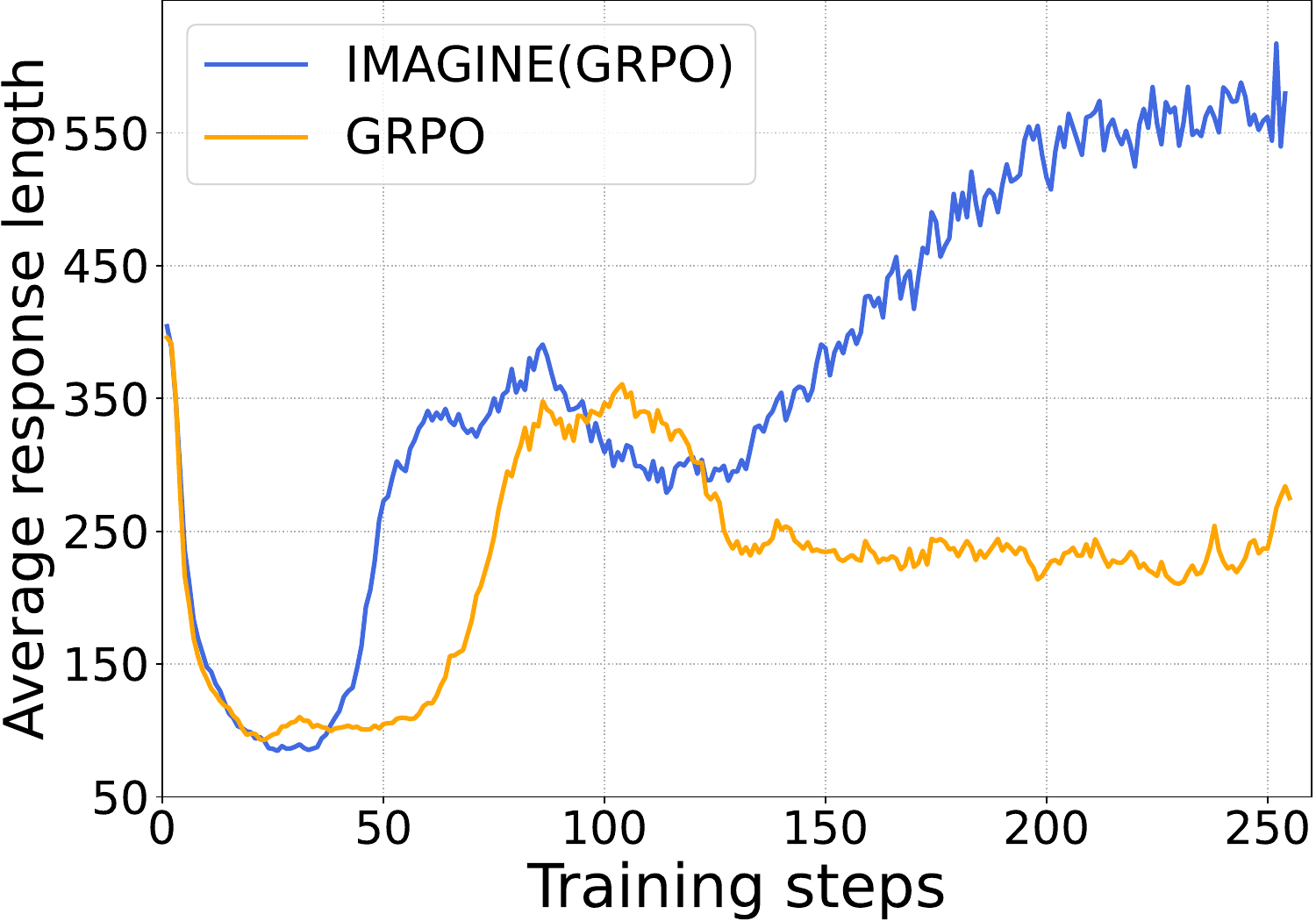}
    \vspace{-2mm}
    \caption{Response length comparison on Countdown-4.}
    \label{fig:relen}
    \vspace{-5mm}
\end{figure}

\textbf{A Case Study} is presented in Figure~\ref{fig:case} to illustrate the significance of \name. Specifically, vanilla GRPO-trained LLM exhibits logical errors (e.g., missing numbers, redundancy in \ding{172}), repetitive reasoning patterns (\ding{173}) due to intensive exploitations without explorations, and failures to solve tasks (\ding{174}) or apply critical operations like multiplication (\ding{175}). With the guidance of the exploration rewards from \name, LLM trained with \name(GRPO) reduces logical errors, diversifies reasoning paths (\ding{175}), and results in a correct answer (\ding{174}), demonstrating improved reasoning ability. Additionally, as shown in Figure~\ref{fig:relen}, \name increases the average response length during training, stabilizing at a higher level than vanilla RL methods. This suggests that \name’s exploration strategies enable LLMs to learn more complex reasoning processes with longer CoTs through considering diverse reasoning paths in responses, leading to performance gains. In the following subsections, we will systematically introduce \name through detailing its key components.

\subsubsection{Trajectory-aware Exploration Rewards}\label{sec:tra}
\name provides intrinsic exploration rewards for LLM reasoning using a simple novelty-by-prediction mechanism. This approach is computationally efficient and does not require task-specific prior knowledge. It maintains two lightweight networks: a randomly initialized and frozen target network $f_{\theta_T}$ and a trainable predictor $f_{\theta_P}$ (with the same architecture). The predictor is trained to match the target on sequences observed during training:
\begin{equation}
\mathcal{L}(\boldsymbol{s}) = \|f_{\theta_P}(\boldsymbol{s}) - f_{\theta_T}(\boldsymbol{s})\|_2^2
\end{equation}
where $\boldsymbol{s}$ denotes the current ``state'' in the RL paradigm. Since the two networks share the same architecture, this loss is guaranteed to converge~\cite{burda2018exploration}. In the autoregressive LLM generation setting, it corresponds to the current generated question-response prefixes $[q,o_{\leq t}]$. Intuitively, if a sequence has been visited frequently, the predictor can fit the target well, and the error becomes small; if it is rarely visited, the error remains large. Therefore, as illustrated in previous studies~\cite{burda2018exploration}, the prediction error could be applied as an intrinsic novelty signal:
\begin{equation}
R^{nov}(\boldsymbol{s}) = \frac{\|f_{\theta_P}(\boldsymbol{s}) - f_{\theta_T}(\boldsymbol{s})\|_2^2}{\operatorname{std}(\|f_{\theta_P}(\boldsymbol{s}) - f_{\theta_T}(\boldsymbol{s})\|_2^2)}.
\end{equation}
The denominator uses a moving estimate of the standard deviation during training to keep the reward scale stable, and remains fixed during evaluation for determinism.

\begin{figure}[t]
% \vspace{-1mm}
    \centering
    \includegraphics[width=\linewidth]{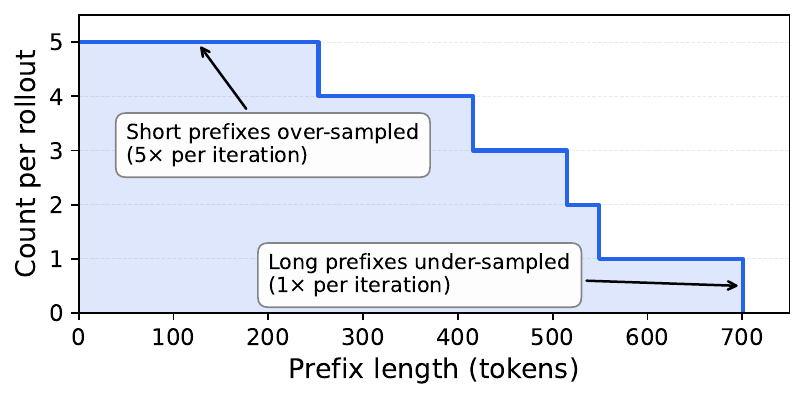}
    \vspace{-5mm}
    \caption{Input statistics for predictor update.}
    \label{fig:token}
    \vspace{-4mm}
\end{figure}

\textbf{Unresolved Challenges:} However, through careful empirical analysis, we identify that naively applying token-level novelty signals to LLMs results in severe performance degradation and computational inefficiency. In autoregressive LLM reasoning, when states are defined at the token level (i.e., $\boldsymbol{s}_t=[q, o_{\leq t}]$), which represents the standard paradigm for autoregressive generation, a single sampled response $o$ would produce a total number of $|o|$ states $\{\boldsymbol{s}_t\}_{t=1}^{|o|}$ of varying lengths for predictor update. This formulation introduces two critical practical challenges: (1) \textbf{Dynamic episodic length.} The distribution of prefix lengths exhibits systematic bias: shorter prefixes occur in every sampled response, whereas longer prefixes appear exclusively in extended responses. This leads to severe under-sampling of longer sequences, as illustrated in Figure~\ref{fig:token}. (2) \textbf{Computational overload.} Token-level state processing necessitates $O(|o|)$ forward passes and predictor updates per sampled response, resulting in prohibitive computational costs that scale quadratically with the number of rollouts.

These phenomena are evident in Figure~\ref{fig:token}: for 5 rollout responses of one question with lengths $[253, 416, 515, 549, 701]$, token-level state processing already yields $253+416+515+549+701=2434$ inputs for predictor update, dominated by short prefixes.

To address these challenges, we propose trajectory-aware exploration rewards operating at the sequence level. Specifically, we keep the same dual-network design but process the complete response sequence $(q,o)$ as a single input instead of using each $[q, o_{\leq t}]$:
\begin{equation}\label{equ:r1}
\left\{
\begin{aligned}
&\mathcal{L}(q, o) = \|f_{\theta_P}(q, o) - f_{\theta_T}(q, o)\|_2^2\\
&R^\star_1(q, o) =
\frac{\|f_{\theta_P}(q, o) - f_{\theta_T}(q, o)\|_2^2}
{\operatorname{std}(\|f_{\theta_P}(q, o) - f_{\theta_T}(q, o)\|_2^2)}
\end{aligned}
\right.
\end{equation}
Here, $\mathcal{L}(q,o)$ is used to update $f_{\theta_P}$, while $R^\star_1(q,o)$ provides a single exploration reward for each complete response. This design eliminates token-level bias through uniform sequence treatment and reduces computation from $O(|o|)$ to $O(1)$ per training sequence, while remaining consistent with RLVR outcome-based reward paradigms.

% \begin{figure}[t]
%     \centering
%     \subfigure[Response length.]{
%         \label{fig:relen}
%     \includegraphics[width=0.7\linewidth]{fig/responselen.pdf}}
%     \subfigure[RND input statistics.]{
%         \label{fig:token}
        
%     \includegraphics[width=0.7\linewidth]{fig/token.png}}
%     \vspace{-4mm}
%     \caption{Training statistics on Countdown-4~\citep{gandhi2024stream}.}
%     \label{fig:3train}
% \end{figure}

\subsubsection{Error-conditioned Reward Allocation}\label{sec:scaling}

Despite the benefits of trajectory-aware exploration rewards, the large action space in LLM reasoning still introduces training challenges. Naive exploration strategies become infeasible due to unstable and sparse rewards and suboptimal exploration efficiency. Moreover, uniform exploration rewards across correct and incorrect samples reduce exploration efficiency and induce instability in correct trajectories. To enable more effective exploration, we prioritize samples with residual errors to facilitate diverse discoveries while efficiently improving overall performance. Thus, \name employs an error-conditioned reward allocation mechanism across sampling sequences. Specifically, a selection function $I_{a\neq o}$ is designed to allocate exploration rewards exclusively to incorrect samples:
\begin{equation}\label{equ:r3}
R^\star(q, o) = I_{a\neq o} \cdot R^\star_1(q, o)
\end{equation}
where $a$ is the ground-truth answer. This conditioning directs exploration resources toward samples with residual errors, enhancing reward utilization while stabilizing predictions for correct samples.

\subsubsection{Advantage-preserving Integration}
Since the intensity of the exploration behavior varies at different stages of training, directly combining exploration rewards with outcome-based rewards creates conflicting learning signals across training stages. For PPO, this injects noise into value function estimation. For GRPO, exploration rewards may invert originally positive advantage signs after group normalization, which is a critical issue since exploration rewards should never penalize samples. To resolve this conflict and ensure seamless
integration with established RL algorithms like PPO and GRPO, \name applies exploration rewards \textbf{after} advantage computation and possible value estimation through:
\begin{equation}\label{equ:adv}
\hat{A}_{new} = \hat{A}_{old} + R^\star(q, o)
\end{equation}
where $\hat{A}_{\text{old}}$ denotes the original advantage from outcome-based rewards. For PPO, we apply the exploration reward to the last advantage token. For GRPO, we apply it to each advantage in a rollout. This design preserves two key properties: (1) The outcome-based advantage maintains its original mean and variance statistics, ensuring stable policy updates; (2) Exploration rewards $R^\star(q, o)$ provide trajectory-level incentive-based guidance without distorting value estimation. The decoupled formulation prevents exploration rewards from conflicting with outcome-based advantage normalization in GRPO or perturbing value function update in PPO, enabling harmonious integration of both reward components.

By applying trajectory-aware exploration rewards with error-conditioned reward allocation and advantage-preserving integration, \name efficiently enhances LLM reasoning by guiding the model toward diverse possible responses during training, while significantly improving overall reasoning ability. Algorithmically, \name employs lightweight networks that operate independently of the LLM, enabling efficient updates and modular reward computation. This design allows \name to integrate seamlessly with standard RLVR algorithms such as PPO and GRPO, consistently yielding performance improvements. The complete algorithm is detailed in Appendix~\ref{sec:alg}.
\section{Experiments}

Here we evaluate \name on three datasets to investigate the following research questions:

\begin{itemize}[leftmargin=*]
\item \textbf{Q1}: How does  \name enhance LLM reasoning performance?
\item \textbf{Q2}: How does \name generalize across models and data distributions?
\end{itemize}

In the following subsections, we begin by outlining the experimental setup, followed by a systematic results analysis addressing our core questions. In addition, due to limited space, Appendix~\ref{sec:detailcase} also provides an expanded case study for Figure~\ref{fig:case}, while implementation specifics--including pseudo code (Appendix~\ref{sec:alg}), data statistics (Appendix~\ref{sec:data}), and configuration details (Appendix~\ref{sec:imp})--are deferred for completeness. Beyond core experiments, Appendix~\ref{sec:naive} validates \name's superiority over basic exploration techniques. We further include an ablation study (Appendix~\ref{sec:ablation}) quantifying submodule contributions, a sensitivity analysis (Appendix~\ref{sec:paramsen}) examining hyperparameter impacts on performance, and a group size analysis in Appendix~\ref{sec:group} for robustness validation.

\subsection{Experimental Setup}

\subsubsection{Dataset}
Following previous studies~\cite{yu2025dapo,tinyzero}, we conduct experiments on three public datasets, i.e., GSM8K\citep{cobbe2021gsm8k}, Countdown-34 and its harder version Countdown-4\citep{gandhi2024stream} to validate \name's effectiveness against vanilla PPO and GRPO algorithms. For computational efficiency, we use a subset of the complete dataset of Countdown-34 and Countdown-4 for training. The detailed dataset statistic is shown in Appendix~\ref{sec:data}. 
Additionally, we also provide detailed experiments on the AIME 2024 dataset between GRPO and \name(GRPO) to further validate \name's effectiveness with multiple outputs in Table~\ref{tab:aime}.

\begin{table*}[t]
% \renewcommand\arraystretch{1.5}
% \vspace{-4mm}
  \centering
  \caption{Average accuracy. \name(PPO)/(GRPO) denotes the implementation of \name on PPO/GRPO. ``Improve'' indicates relative improvement.}
  \vspace{-2mm}
  \scalebox{0.8}{
    \begin{tabular}{ccccccc}
    \toprule
    Dataset & PPO   & \name(PPO) & \textbf{Improve} & GRPO  & \name(GRPO) & \textbf{Improve} \\
    \midrule
    GSM8K & 0.8051$\pm$0.0057 & 0.8169$\pm$0.0042 & \textbf{1.47\%} & 0.8082$\pm$0.0102 & 0.8251$\pm$0.0121 & \textbf{2.09\%} \\
    Countdown-34 & 0.5526$\pm$0.0332 & 0.5924$\pm$0.0273 & \textbf{7.2\%} & 0.6711$\pm$0.0213 & 0.7132$\pm$0.0199 & \textbf{6.27\%} \\
    Countdown-4 & 0.3307$\pm$0.0242 & 0.3812$\pm$0.0281 & \textbf{15.27\%} & 0.3872$\pm$0.0452 & 0.4739$\pm$0.0574 & \textbf{22.39\%} \\
    \bottomrule
    \end{tabular}%
    }
  \label{tab:mainresult}
  \vspace{-3mm}
\end{table*}%

\begin{figure*}[t]
    \centering

    \subfigure[GSM8K with PPO.]{
        \label{fig:gsppo}
    \includegraphics[width=0.27\linewidth]{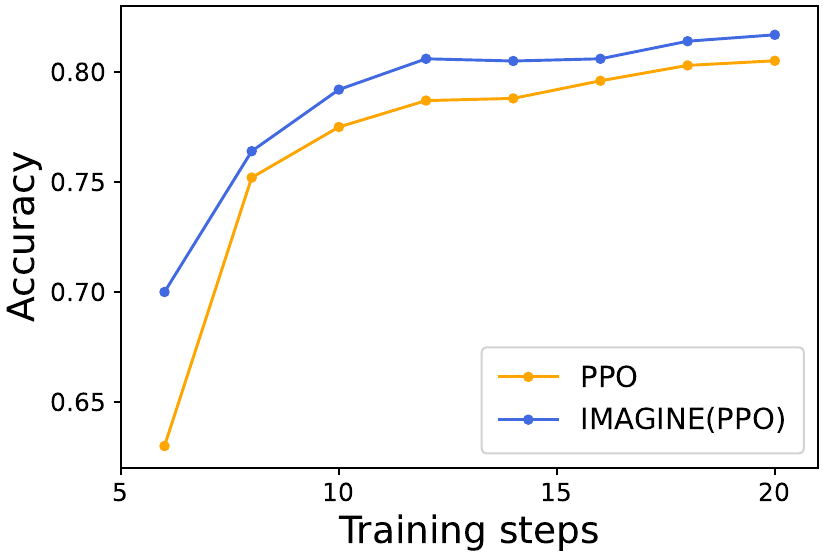}}
    \subfigure[Countdown-34 with PPO.]{
        \label{fig:c34ppo}
    \includegraphics[width=0.27\linewidth]{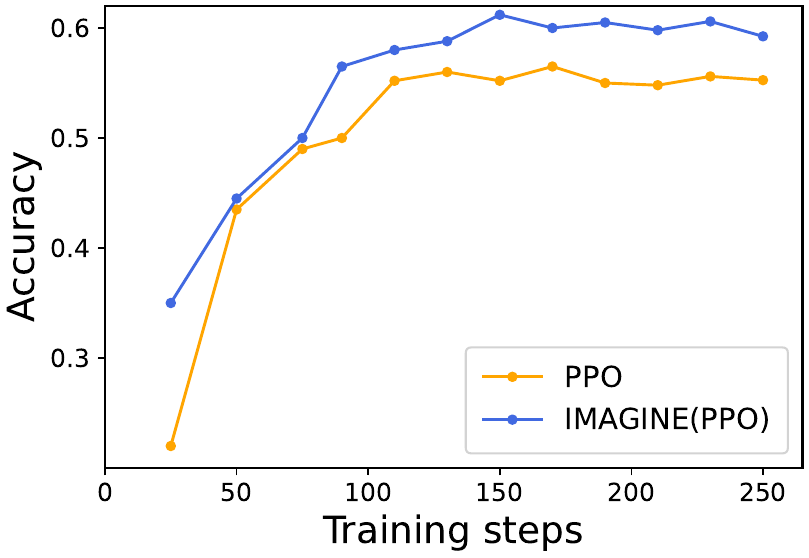}}
    \subfigure[Countdown-4 with PPO.]{
        \label{fig:c4ppo}
    \includegraphics[width=0.27\linewidth]{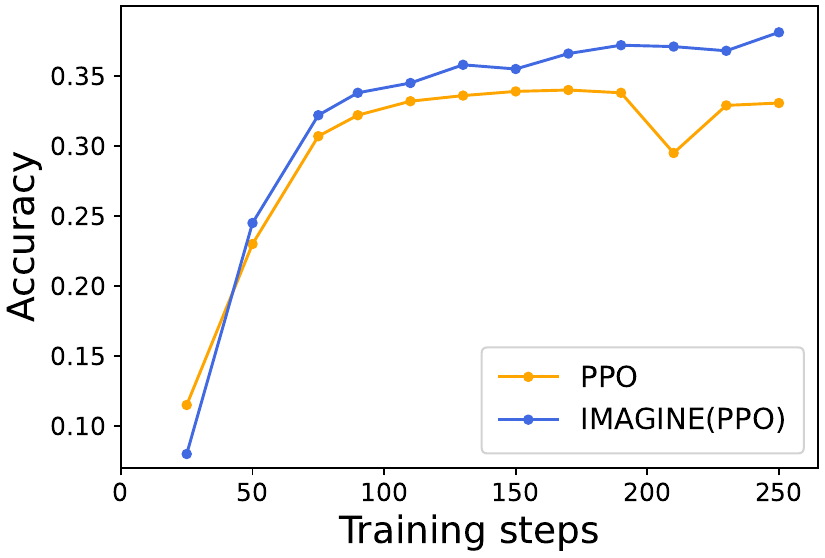}}
    \vspace{-2mm}

    \subfigure[GSM8K with GRPO.]{
        \label{fig:gsgrpo}
    \includegraphics[width=0.27\linewidth]{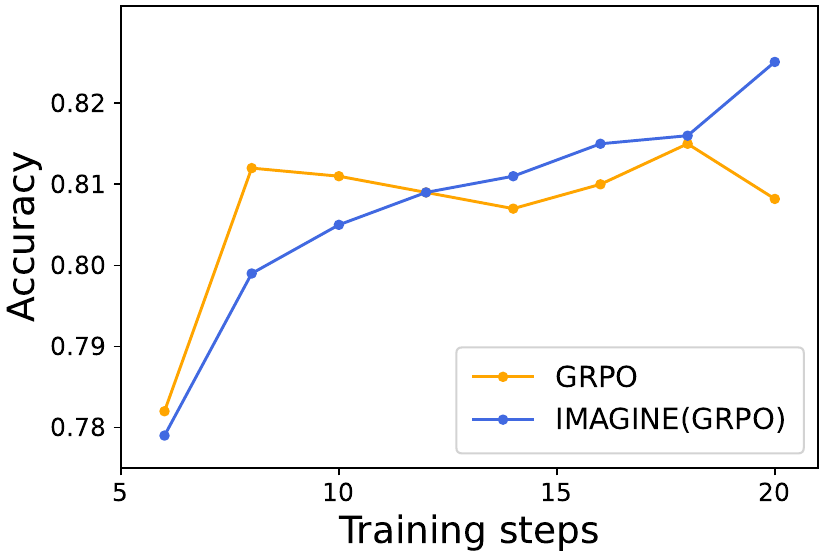}}
    \subfigure[Countdown-34 with GRPO.]{
        \label{fig:c34grpo}
    \includegraphics[width=0.27\linewidth]{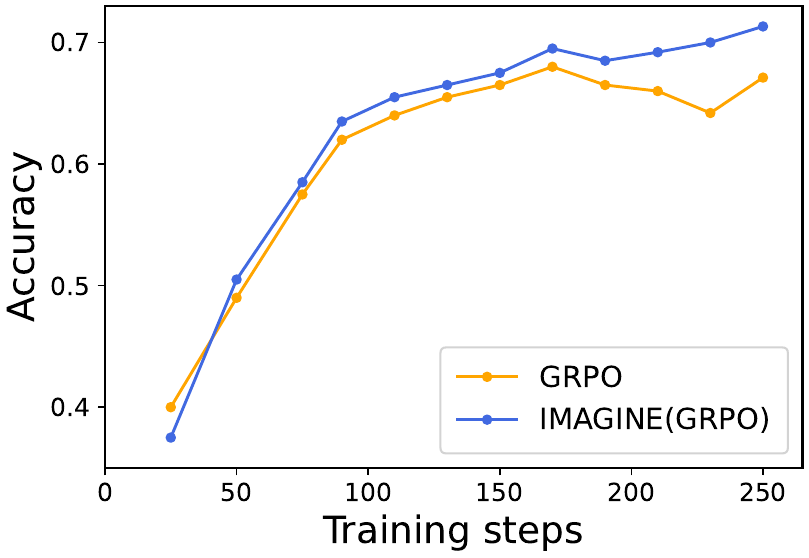}}
    \subfigure[Countdown-4 with GRPO.]{
        \label{fig:c4grpo}
    \includegraphics[width=0.27\linewidth]{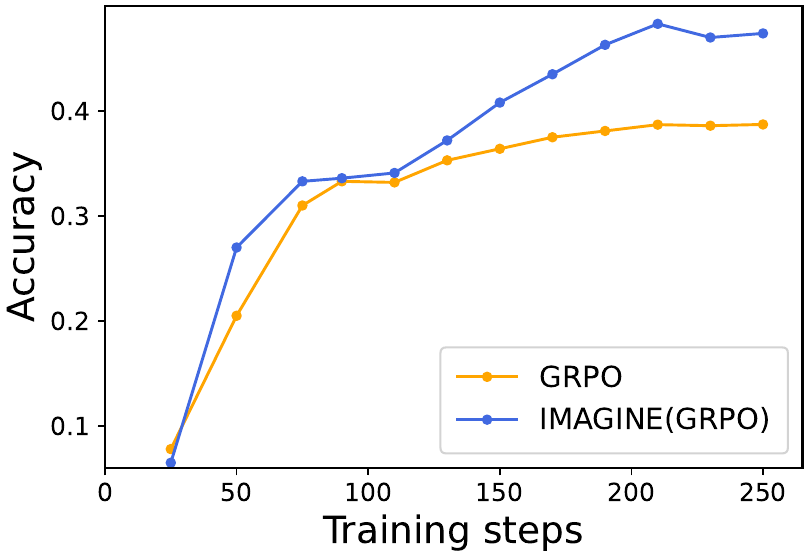}}
    \vspace{-2mm}
    \caption{Evaluation accuracies in some of the training trajectories.}
    \vspace{-3mm}
    \label{fig:c4}
\end{figure*}

\subsubsection{Evaluation Protocol}\label{sec:evaluate}

\textbf{Environment and LLM backbone:} The experiment environment is built on the TinyZero~\footnote{https://github.com/Jiayi-Pan/TinyZero}\citep{tinyzero} and verl~\footnote{https://github.com/volcengine/verl} framework with Qwen2.5-3B~\citep{qwen2, qwen2.5} as base model. We also provide generalization experiments on deepseek-7b in Section~\ref{sec:deepseek}. Similar to DAPO~\citep{yu2025dapo}, we exclude the KL penalty from RL algorithms after a detailed analysis in Appendix~\ref{sec:naive}.

\textbf{Network structure of \name:} The predictor and target networks in \name share the same architecture, which is dataset-agnostic and LLM-independent. For computational efficiency, we simply apply a lightweight predictor and target network structure for \name (embedding size$=16$, three FFN layers with neurons $[16,8,1]$), ensuring negligible additional training time.
To make a fairer comparison, we apply a set of fixed training hyperparameters to all datasets. 

\textbf{Evaluation metric:} For evaluation, we report the average accuracy on five runs for each experiment. 

Implementation details can be found in Appendix~\ref{sec:imp} to further elaborate on the training settings.

\subsection{Main Result (Q1)}

The main results are presented in Table~\ref{tab:mainresult}. Figure~\ref{fig:c4} highlights key training details via some training trajectories, showcasing the performance improvements achieved during training. From Table~\ref{tab:mainresult} and Figure~\ref{fig:c4} we could conclude that:
\begin{itemize}[leftmargin=*]
    \item GRPO outperforms PPO on all datasets, illustrating superior training effects brought by group-wise sampling and reward normalization. Both \name(PPO) and \name(GRPO) outperform standard RL approaches, demonstrating that \name effectively guides and enhances the updates of LLM reasoning via exploration behavior during the RL training process. By encouraging the exploration of new responses rather than merely fitting to outcome-based reward through sampling behaviors, \name enables LLMs to explore more diverse potential inference paths during training and avoids LLMs getting trapped in local optimal solutions.
    
    \item A key challenge in improving LLM reasoning lies in handling difficult samples, which typically yield near-zero rewards that hinder parameter updates. 
    Our experiments reveal an interesting pattern: \name's improvements are most significant on Countdown-4, followed by Countdown-34 and GSM8K. This progression aligns with the relative difficulty levels of these datasets, where Countdown-4 \textgreater{} Countdown-34 \textgreater{} GSM8K. The results suggest that \name's dense exploration rewards specifically help models overcome learning barriers in challenging samples. This phenomenon represents a further contribution of our study: it originates from RLVR’s core distinction from traditional RL--leveraging a pre-trained LLM instead of a randomly initialized model. Consequently, exploration yields significantly greater benefits when addressing complex problems.
    
    \item \name achieves greater performance gains with GRPO than with PPO. This advantage stems from GRPO's rollout mechanism, which enables \name to generate varied exploration rewards for multiple responses to the same question. Such a design promotes broader exploration of potential solution paths compared to single-response PPO updates.

    % \item Results in Table~\ref{tab:aime} demonstrate two key findings: First, LLM accuracy improves consistently with increasing response counts, confirming that aggregating multiple outputs enhances model performance. Second, \name achieves approximately 20\% improvement across all metrics compared to GRPO baselines. This validates \name's effectiveness in solving complex problems while efficiently leveraging multiple responses.
\end{itemize}

% Table generated by Excel2LaTeX from sheet 'Sheet1'
\begin{table}[t]
  \centering
  \vspace{-3mm}
  \caption{Experimental result on AIME 2024.}
  \vspace{-2mm}
  \resizebox{\linewidth}{!}{
    \begin{tabular}{cccccc}
    \toprule
    Method & Pass@1 & Pass@8 & Pass@32 & Avg@8 & Avg@32 \\
    \midrule
    GRPO  & 0.3   & 0.3   & 0.3333 & 0.3   & 0.3073 \\
    \name(GRPO) & 0.3667 & 0.3667 & 0.4   & 0.3667 & 0.3667 \\
    Improve & 22.23\% & 22.23\% & 20.01\% & 22.23\% & 19.33\% \\
    \bottomrule
    \end{tabular}%
    }
    \vspace{-2mm}
  \label{tab:aime}%
\end{table}%

\subsection{Analysis on Multiple Generations (Q1)}\label{sec:aime}
In this section, we validate \name's efficacy on the AIME 2024 benchmark between GRPO and \name(GRPO) following previous studies~\cite{yu2025dapo,liu2025understanding}. For each question, the LLM generates multiple answers to enable more rigorous performance evaluation~\citep{yu2025dapo}. Results in Table~\ref{tab:aime} demonstrate two key findings: First, LLM accuracy improves consistently with increasing response counts, confirming that aggregating multiple outputs enhances model performance. Second, \name achieves approximately 20\% improvement across all metrics compared to GRPO baselines. This validates \name's effectiveness in solving complex problems while efficiently leveraging multiple responses.

\subsection{Generalization Analysis (Q2)}\label{sec:generalization}
In this section, we conduct experimental analysis to verify the generalization ability of \name from both the model and dataset perspectives.

\subsubsection{Base Model Generalization Analysis}\label{sec:deepseek}
Here, we verify the effectiveness of \name on another commonly used base model, deepseek-7b, with GRPO on the GSM8K dataset to validate \name's generalization on different base models. From the result in Table~\ref{tab:basemodel}, we could conclude that \name can be applied to different base models, including Qwen and DeepSeek, and provide stable performance gains for their reasoning process.

% Table generated by Excel2LaTeX from sheet 'deepseek'
\begin{table}[h]
  \centering
  \caption{Experiments with deepseek-7b base model on GSM8K.}
  \vspace{-2mm}
  \scalebox{0.9}{
    \begin{tabular}{ccc}
    \toprule
    Dataset & GRPO  & \name(GRPO) \\
    \midrule
    GSM8K & 0.6535$\pm$0.0049 & 0.6641$\pm$0.0052 \\
    \bottomrule
    \end{tabular}%
    }
    \vspace{-3mm}
  \label{tab:basemodel}%
\end{table}%

\subsubsection{Dataset Generalization Analysis}
This subsection examines \name's impact on generalization capability by evaluating performance on out-of-distribution data. Specifically, we train the LLM exclusively on Countdown-4 and measure performance on the unseen Countdown-34 dataset. As evidenced by Table~\ref{tab:generalization}, \name(GRPO) surpasses GRPO on both Countdown-4 and the held-out Countdown-34 dataset after training. This demonstrates that \name not only enhances optimization efficacy but also preserves - and even strengthens - the model's generalization capacity when exposed to novel data distributions.

% Table generated by Excel2LaTeX from sheet 'Sheet1'
\begin{table}[h]
  \centering
  \caption{Dataset Generalization analysis.}
  \vspace{-2mm}
  \resizebox{\linewidth}{!}{
    \begin{tabular}{cccc}
    \toprule
    Dataset & GRPO  & \name(GRPO) & \textbf{Improve} \\
    \midrule
    Coundown-4 & 0.3831$\pm$0.0482 & 0.4726$\pm$0.0418 & \textbf{23.36\%} \\
    \midrule
    Countdown-34 & 0.3369$\pm$0.0531 & 0.3877$\pm$0.0563 & \textbf{15.08\%} \\
    \bottomrule
    \end{tabular}%
    }
    \vspace{-3mm}
  \label{tab:generalization}%
\end{table}%

% Table generated by Excel2LaTeX from sheet 'time'
\begin{table}[t]
  \centering
  \caption{Training time analysis.}
  \vspace{-2mm}
  \resizebox{\linewidth}{!}{
    \begin{tabular}{ccccc}
    \toprule
    Dataset & PPO   & \name(PPO) & GRPO  & \name(GRPO) \\
    \midrule
    GSM8K & 0.70h & 0.73h & 0.86h & 0.88h \\
    Countdown-34 & 7.43h & 7.81h & 12.13h & 14.55h \\
    Countdown-4 & 6.97h & 7.20h & 11.69h & 13.03h \\
    \bottomrule
    \end{tabular}%
    }
    \vspace{-2mm}
  \label{tab:time}%
\end{table}%

\section{Training Time Analysis}\label{sec:time}
Here we summarize the training time for experiments in Table~\ref{tab:mainresult} to examine the additional cost in terms of time that \name incurs. The results in Table~\ref{tab:time} demonstrate that \name introduces only minimal training time overhead versus baselines without it. This efficiency stems from the predictor and target networks requiring small parameter counts--they comprise several compact linear layers. These networks compute rewards via output differences to determine trajectory visit frequency during training, without needing complex semantic understanding. Critically, since the exploration module operates solely during training, \name has zero inference overhead. This confirms \name's superiority in delivering performance gains with negligible training time impact.

\section{Exploration Analysis}
In this section, we quantified the exploration effect of GRPO and \name(GRPO) during training to analyze the promoting role of \name on exploration behavior. Specifically, as Section~\ref{sec:tra} and Equation~\eqref{equ:r1} have stated, the convergence rate of the loss $\mathcal{L}(q, o)$ provided by \name represents the frequency of occurrence of similar trajectories during the training process. The higher the frequency of similar trajectories, the faster $\mathcal{L}(q, o)$ decreases on this part of the trajectory, and at the same time, the smaller the average value of $\mathcal{L}(q, o)$ within such a batch. Therefore, the decrease speed of this value can be used as a criterion to judge the diversity of trajectories in the training stage.

% \begin{minipage}{\textwidth}

% \begin{minipage}[h]{0.48\textwidth}
% \makeatletter\def\@captype{figure}
% \centering  
% \includegraphics[width=1\linewidth]{fig/ablation.pdf}  
% \vspace{-3mm}
% \caption{Exploration analysis on Countdown-4.} 

% \label{fig:expana}  
% \end{minipage}
% \begin{minipage}[h]{0.48\textwidth}
% \makeatletter\def\@captype{table}
% \caption{Scalability analysis on Countdown-34.}
% \centering
% \begin{tabular}{ccc}
%     \toprule
%     Model & GRPO  & \name(GRPO) \\
%     \midrule
%     Qwen2.5-0.5B &   0.0033    & 0.0107 \\
%     Qwen2.5-1.5B &       &  \\
%     Qwen2.5-3B &    0.6711   & 0.7132 \\
%     \bottomrule
%     \end{tabular}%
% \label{tab:sca}
% \end{minipage}
% \end{minipage}

\begin{figure}[h]  
    \centering  
    \includegraphics[width=0.9\linewidth]{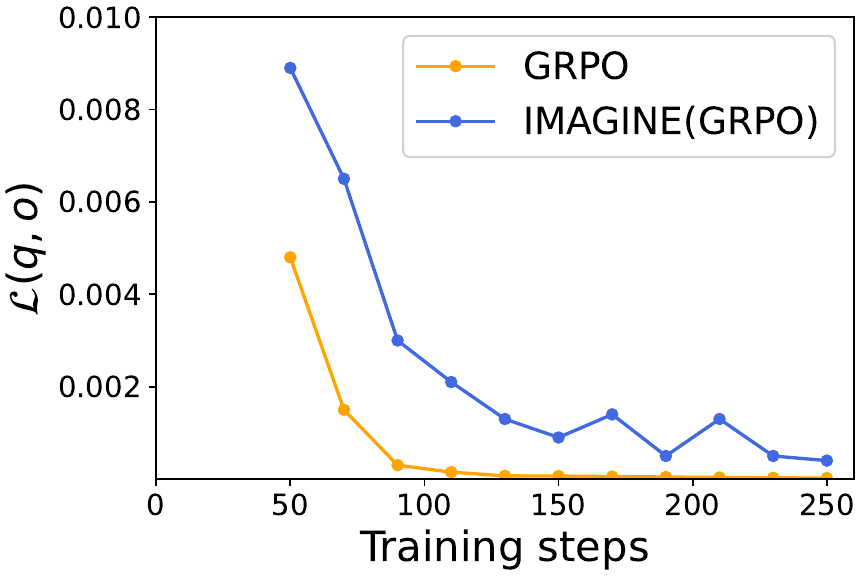}  
    \vspace{-2mm}
    \caption{Exploration analysis on Countdown-4.} 
    \vspace{-2mm}
    \label{fig:expana}  
    
\end{figure}  

As in Figure~\ref{fig:expana}, the decrease speed of $\mathcal{L}(q, o)$ on GRPO is significantly faster than that of \name(GRPO). This shows that the trajectories generated during training with \name(GRPO) are more diverse than those with GRPO.

\section{Scalability Analysis}
To further test the scalability of \name on different model sizes, we experiment the performance of \name(GRPO) on Countdown-34 with different base model sizes: Qwen2.5- 0.5B/1.5B/3B. The results are shown in Table~\ref{tab:sca}. From the table, we can conclude that \name can be applied to models of different sizes, bringing stable performance improvements.

% Table generated by Excel2LaTeX from sheet 'Sheet1'
\begin{table}[t]
  \centering
  \caption{Scalability analysis on Countdown-34.}
  \vspace{-2mm}
  \scalebox{0.9}{
    \begin{tabular}{ccc}
    \toprule
    Model & GRPO  & \name(GRPO) \\
    \midrule
    Qwen2.5-0.5B &   0.0033    & 0.0107 \\
    Qwen2.5-1.5B &  0.4532     & 0.5068 \\
    Qwen2.5-3B &    0.6711   & 0.7132 \\
    \bottomrule
    \end{tabular}%
    }
  \label{tab:sca}%
  \vspace{-3mm}
\end{table}%

\section{Compatibility Analysis}
Owing to \name’s lightweight structure and strong compatibility, it can be applied directly to PPO and GRPO, and it can also be combined with recent advanced RL variants such as DAPO~\cite{yu2025dapo}, potentially facilitating exploration and improving performance. In this section, we further examine \name’s compatibility with GRPO-based techniques. Accordingly, we evaluate the combination of DAPO and \name on Countdown-4. As shown in Table~\ref{tab:compat}, the combined method consistently outperforms either component alone, suggesting that \name is complementary to DAPO and supporting its applicability across different training variants.

\begin{table}[h]
\centering
\vspace{-2mm}
\caption{Compatibility analysis.}
\vspace{-2mm}
\centering
\scalebox{0.9}{
\begin{tabular}{cc}
    \toprule
    Method & Accuracy \\
    \midrule
    DAPO  & 0.4691 \\
    \name(DAPO) & 0.4924 \\
    \bottomrule
    \end{tabular}%
    }
\label{tab:compat}
\vspace{-2mm}
\end{table}

\section{Related Works}
\subsection{Reinforcement Learning for LLMs}

Reinforcement learning has evolved from foundational approaches like PPO~\citep{schulman2017proximal} to advanced frameworks such as GRPO~\citep{shao2024deepseekmath}. PPO achieves stability in policy updates through clipped objectives and trust region constraints, effectively aligning models for dialogue systems and code generation~\citep{wang2024enhancing,shojaee2023execution}. 
Though joint policy-value optimization introduces computational overhead~\citep{shao2024deepseekmath}.
% Its architecture treats token positions in reasoning trajectories of LLM as distinct states for advantage estimation~\citep{hu2025openreasonerzeroopensourceapproach}, though joint policy-value optimization introduces computational overhead~\citep{shao2024deepseekmath}.
GRPO~\citep{shao2024deepseekmath} mitigates these limitations through trajectory-level sampling and group-wise advantage normalization. By generating multiple responses per input and standardizing rewards across trajectory groups, GRPO reduces policy update bias while maintaining efficiency. 
% This sequence-level evaluation mechanism decouples advantage estimation from computational complexity, proving particularly effective in multi-step reasoning scenarios. 
Recent implementations such as DAPO~\citep{yu2025dapo} and Dr. GRPO~\citep{liu2025understanding} further showcase GRPO's scalability through detailed refinement of reasoning objectives, though exploration efficiency remains constrained by static and sparse reward structures~\citep{dou2025improving,gao2024designing}. Differently, \name enhances LLM reasoning through providing dense, stable, and efficient exploration rewards in RL optimization process. These rewards enable intrinsic motivation-guided exploration over diverse response trajectories during training, systematically improving reasoning capabilities.

\subsection{Improving Reasoning Ability of LLMs}\label{sec:related}  
Recent advances in LLM reasoning focus on three key paradigms: \textbf{Pre-training augmentation} improves foundational reasoning by exposing models to curated datasets and synthetic traces. Models like Qwen~\citep{yang2024qwen2} and Llama~\citep{grattafiori2024llama} show significant gains through domain-specific data scaling, though this requires heavy computational resources and careful data curation. \textbf{Prompt engineering}~\citep{sahoo2024systematic,chen2023unleashing} enhances reasoning via structured inputs. Techniques like Chain-of-Thought~\citep{yu2023towards,chu2023navigate} decompose problems into steps, while self-consistency~\citep{wang2022self} improves outputs through majority voting. Extensions such as multi-agent debates~\citep{liang2023encouraging} and iterative refinement~\citep{madaan2023self} further extend capabilities but are limited by manual design and high computational costs.  \textbf{Algorithmic enhancement} combines inference-time search with RL-based fine-tuning. Methods like Monte Carlo Tree Search~\citep{browne2012survey} and RL approaches (e.g., PPO and GRPO) optimize reasoning policies but often overexploit fixed high-reward paths, limiting exploration of novel solutions. Unlike these methods, \name actively incentivizes exploration of under-optimized paths through exploration rewards, enabling deeper action-space traversal.
\section{Conclusion}  
We introduce \name, a reinforcement learning framework that systematically addresses the challenges of sparse rewards and limited exploration in RL-based LLM finetuning through three methodological advances. First, our trajectory-aware exploration reward employs lightweight networks to provide dense exploration rewards, enabling efficient discovery of high-quality reasoning paths. Second, we propose error-conditioned reward allocation to ensure efficient exploration on challenging samples while intrinsically stabilizing training in large action spaces, ensuring sustained exploration. Third, our advantage-preserving integration decouples exploration incentives from outcome-based policy gradients, resolving fundamental conflicts between intrinsic and extrinsic rewards. Extensive experiments demonstrate that \name significantly enhances reasoning performance when integrated with both PPO and GRPO, establishing its versatility across distinct RL paradigms. 
% By stabilizing the exploration-exploitation dynamics in complex reasoning tasks, our work provides a principled approach for discovering diverse trajectories while maintaining efficiency.

\end{sloppypar}

\nocite{langley00}
\balance
\bibliography{bibfile}
\bibliographystyle{icml2026}

%%%%%%%%%%%%%%%%%%%%%%%%%%%%%%%%%%%%%%%%%%%%%%%%%%%%%%%%%%%%%%%%%%%%%%%%%%%%%%%
%%%%%%%%%%%%%%%%%%%%%%%%%%%%%%%%%%%%%%%%%%%%%%%%%%%%%%%%%%%%%%%%%%%%%%%%%%%%%%%
% APPENDIX
%%%%%%%%%%%%%%%%%%%%%%%%%%%%%%%%%%%%%%%%%%%%%%%%%%%%%%%%%%%%%%%%%%%%%%%%%%%%%%%
%%%%%%%%%%%%%%%%%%%%%%%%%%%%%%%%%%%%%%%%%%%%%%%%%%%%%%%%%%%%%%%%%%%%%%%%%%%%%%%
\newpage
\appendix
\onecolumn
\section{Detailed Case Study}\label{sec:detailcase}
\begin{figure}[h]
% \vspace{-1mm}
    \centering
    \includegraphics[width=0.8\linewidth]{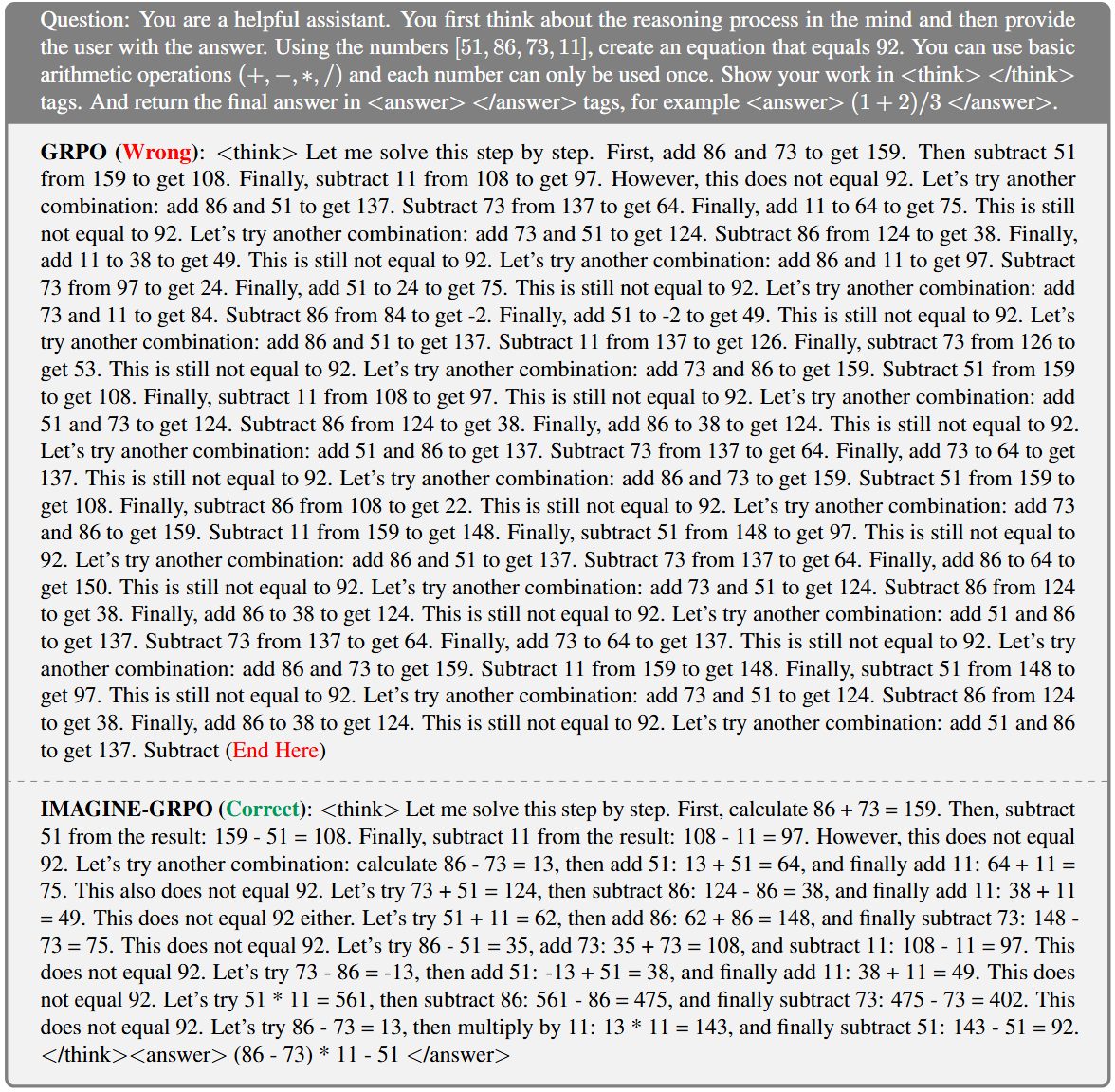}
    % \vspace{-2mm}
    \caption{Detailed case study text.}
    \label{fig:detailcase}
    % \vspace{-3mm}
\end{figure}

We provide the full response texts from the case study (Figure~\ref{fig:case}) in Figure~\ref{fig:detailcase} for reference. Notably, the incorrect responses are longer than the correct ones, a common trend observed across all algorithms examined in this paper. This occurs because a well-trained LLM typically stops generating text once it arrives at the correct answer, whereas incorrect responses involve exploring more possible combinations to find a solution. However, as shown in Figure~\ref{fig:relen}, the overall average response length of \name(GRPO) is longer than that of GRPO. This suggests that LLMs trained with \name(GRPO) develop more complex reasoning abilities, enabling them to approach problems from a broader range of perspectives.

\section{\name Algorithm}\label{sec:alg}  
Through the methodological components described in Section~\ref{sec:methoddetail}, \name implements a reinforcement learning framework with structured exploration guidance to enhance LLM reasoning capabilities. The complete optimization procedure, formalized in Algorithm~\ref{alg:GERO}, operates as follows: Given an input batch $B$ of question-response pairs $(q, o)$ for advantage estimation, we first compute the baseline outcome-based reward signals $R$ and initial advantage estimates $\hat{A}_{\text{old}}$ using conventional outcome-based reward functions and RL advantage estimation methods. Subsequently, each sample is processed through \name's dual networks $(f_{\theta_P}, f_{\theta_T})$ to simultaneously update the policy network $f_{\theta_P}$ and obtain trajectory-aware exploration rewards $R^\star_1(q, o)$. These exploration rewards are then adaptively scaled through our error-conditioned reward allocation approach, yielding $R^\star(q, o)$ that maintains training stability across different optimization phases. Finally, the refined advantages $\hat{A}_{\text{new}}$ are computed through our advantage-preserving integration mechanism that injects $R^\star(q, o)$ into $\hat{A}_{\text{old}}$ without distorting the original policy gradient signals. The resulting $\hat{A}_{\text{new}}$ subsequently drives policy updates through standard RL optimization algorithms like PPO and GRPO, enabling effective LLM optimization while maintaining gradient stability.

\begin{algorithm}[h]
	\caption{\label{alg:GERO} Optimization algorithm of \name}
	\label{alg:1}
	\raggedright
	{\bf Input}: A coming batch $B$ of $(q,o)$ samples for advantage estimation. A fixed randomly initialized network $f_{\theta_T}$, a predictor network $f_{\theta_P}$  with identical architecture as $f_{\theta_T}$.\\
	{\bf Output}: Advantage $\hat{A}_{new}$ for policy update with RL algorithms such as PPO and GRPO.\\
	\begin{algorithmic} [1]
	    \STATE Obtain outcome-based rewards $R$ through outcome-based reward function
        \STATE Obtain outcome-based advantage $\hat{A}_{old}$ via fixed reward functions
        \STATE Update $f_{\theta_P}$ according to loss function $\mathcal{L}$ in Equation~\eqref{equ:r1}
        \STATE Obtain $R^\star_1(q, o)$ via Equation~\eqref{equ:r1}
        \STATE Obtain $R^\star(q, o)$ via Equation~\eqref{equ:r3}
        \STATE Add the exploration reward $R^\star(q, o)$ to $\hat{A}_{old}$ for a new advantage $\hat{A}_{new}$ via Equation~\eqref{equ:adv}
        \STATE return $\hat{A}_{new}$~\\
	\end{algorithmic}
\end{algorithm}

\section{Dataset Statistics}\label{sec:data}
This section briefly introduces the datasets used in this paper. Specifically, GSM8K~\footnotemark[1] is a dataset of 8.5K high-quality linguistically diverse grade school math word problems. Countdown-34~\footnotemark[2] and Countdown-4~\footnotemark[3] are two mathematical datasets that perform combined operations based on several given numbers to obtain a given answer. Among them, the input sample of Countdown-34 contains 3 or 4 numbers, while the input sample of Countdown-4 only contains four numbers, making its average difficulty higher. For computational efficiency, we use a subset of the complete dataset of Countdown-34 and Countdown-4 for training. The detailed dataset statistic is shown in Table~\ref{tab:data}.

% Table generated by Excel2LaTeX from sheet 'Sheet1'
\begin{table}[h]
  \centering
  \caption{Data Statistics.}
    \begin{tabular}{cccc}
    \toprule
    Params & GSM8K & Countdown-34 & Countdown-4 \\
    \midrule
    Training samples & 7,473 & 32,768 & 32,768 \\
    Testing samples & 1,319 & 1,024 & 1,024 \\
    Max prompt length & 256   & 256  & 256 \\
    Max response length & 1,024  & 1,024  & 1,024\\
    \bottomrule
    \end{tabular}%
  \label{tab:data}%
\end{table}%

\section{Implementation Details}\label{sec:imp}
In this paper, we use 4 GPUs for each training loop. Due to variations in dataset sizes and difficulty levels, we train for 250 steps on Countdown-34 and Countdown-4, and 20 steps on GSM8K, using a batch size of 512 to ensure convergence. Unless otherwise specified, all experiments are conducted with fixed hyperparameters for fair comparison. Based on preliminary grid search experiments, the exploration intensity parameter $\alpha$ is set to 0.5, and the exploration attenuation rate $\gamma$ is set to 40, ensuring an averaged optimal performance of \name across all datasets. For GRPO, the rollout group size is set to 5. For outcome-based rewards, we adopt the same reward function as TinyZero~\footnotemark[4], defined as:

\begin{equation}\label{equ:outcomereward}
R(o, a) = \begin{cases}1.0, & a == o \\ 0.1, & \text{correct format reward} \\ 0.0, & \text{otherwise} \end{cases}
\end{equation}
where $a$ is the ground truth answer for response $o$. To ensure fair evaluation, we report the average accuracy over five experiments, rather than evaluation scores, to eliminate potential gains from format rewards during evaluation.

\section{Comparison with Basic Exploration Techniques}\label{sec:naive} 
Beyond \name, researchers typically control LLM exploration through two basic techniques: (1) adjusting the temperature parameter $Temp$ to influence output diversity by reshaping token probabilities, and (2) modifying the KL penalty coefficient $\beta$ to regulate how strictly the policy adheres to its original behavior during RL updates. We evaluate these approaches using GRPO with varying KL and temperature coefficients, testing whether performance declines when deviating from their optimal values ($\beta=0.0$, $Temp=1.0$) used by default in this paper. 

Our experiments in Table~\ref{tab:naive} reveal that increasing the KL coefficient (which tightens constraints on policy updates) and reducing the temperature (which introduces limited randomness in training) both degrade reasoning performance--the former limits exploratory updates by over-anchoring to the initial policy, while the latter disrupts sample diversity in training. This validates our baseline configuration with $\beta=0.0$ and $Temp=1.0$ for all RL methods in this paper. Moreover, by introducing exploration rewards that actively guide the model toward novel reasoning paths, \name achieves superior performance. This demonstrates that \name’s trajectory-aware exploration rewards complement rather than conflict with basic exploration mechanisms, providing structured guidance for discovering novel responses while maintaining training stability.

% Table generated by Excel2LaTeX from sheet 'Sheet1'
\begin{table}[h]
  \centering
  \caption{Comparison with Other Naive Exploration Methods. ``$\beta$'' and ``Temperature'' indicate the KL penalty and LLM temperature coefficients.}
    \begin{tabular}{lc}
    \toprule
    Model & Countdown-4 \\
    \midrule
    GRPO ($\beta=0.0$, $Temp=1.0$)  & 0.3872$\pm$0.0452 \\
    \hdashline[3pt/2pt]
    \textbackslash{}w $\beta=0.001$ & 0.3672$\pm$0.0412 \\
    \textbackslash{}w $\beta=0.01$ & 0.1778$\pm$0.0289 \\
    \textbackslash{}w $Temp=0.3$ &  0.2263$\pm$0.0272\\
    \textbackslash{}w $Temp=0.6$ &  0.2869$\pm$0.0337\\
    \hdashline[3pt/2pt]
    \name(GRPO) ($\beta=0.0$, $Temp=1.0$) & 0.4739$\pm$0.0574 \\
    \bottomrule
    \end{tabular}%
  \label{tab:naive}%
\end{table}%

\section{Ablation Study}\label{sec:ablation}
The results from Table~\ref{tab:ablation} highlight the progressive improvements contributed by each component of our proposed approach, \name. Each enhancement effectively addresses core challenges in policy optimization for LLMs, as reflected by the corresponding increase in accuracy at every stage.

% Table generated by Excel2LaTeX from sheet 'Sheet1'
\begin{table}[h]
\vspace{-3mm}
  \centering
  \caption{Ablation study on Countdown-34.}
  \vspace{-3mm}
    \begin{tabular}{lc}
    \toprule
    Model & Accuracy \\
    \midrule
    GRPO  & 0.6711$\pm$0.0213 \\
    $+$Trajectory-aware Exploration Rewards & 0.6939$\pm$0.0217 \\
    $+$Error-conditioned Reward Allocation & 0.7065$\pm$0.0173 \\
    $+$Advantage-Preserving Reward Integration (\name(GRPO)) & 0.7132$\pm$0.0199 \\
    \bottomrule
    \end{tabular}%
  \label{tab:ablation}%
  \vspace{-2mm}
\end{table}%

From the table, we could conclude that: (1) Incorporating trajectory-aware exploration rewards leads to a notable improvement in reasoning performance. This demonstrates that enabling the model to explore diverse potential responses during training promotes a richer learning process. By encouraging the discovery of alternative reasoning paths rather than converging prematurely on a limited set of high-reward responses, the model acquires more nuanced reasoning capabilities, as evidenced by the marked accuracy gain. (2) Adding the error-conditioned reward allocation mechanism further enhances model performance. 
By applying error-conditioned reward allocation, \name conducts efficient exploration with adaptive balance on exploration and exploitation during policy updates, ensuring stable and effective policy optimization. (3) Finally, the advantage-preserving integration ensures seamless compatibility with RL algorithms, such as PPO and GRPO. By preserving the statistical integrity of outcome-based advantage distributions while incorporating exploratory signals, \name achieves a robust balance between exploiting known strategies and exploring new areas, thus enhancing generalization capabilities. With all three components, \name(GRPO) demonstrates the highest improvement in accuracy. 

Overall, the cumulative contributions of these techniques enable \name to outperform the baseline GRPO model significantly, achieving a relative accuracy improvement of +6.27\%. These results validate the effectiveness of our proposed components in addressing the unique challenges of training LLMs for complex reasoning tasks.

\section{Parameter Sensitivity Analysis}\label{sec:paramsen}
\begin{figure}[h]  
    \centering  
    \includegraphics[width=0.6\linewidth]{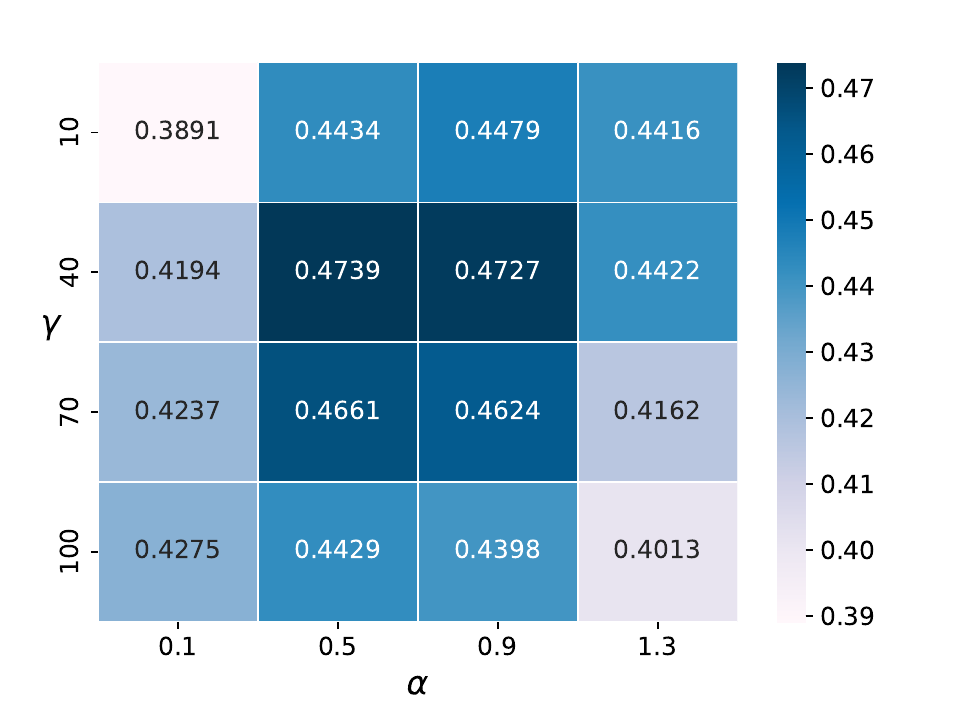}  
    \caption{Sensitivity of \name to hyperparameters $\alpha$ and $\gamma$ on Countdown-4 for \name(GRPO). Optimal performance occurs at $\alpha=0.5$, $\gamma=40$, adopted as default settings.}  
    \label{fig:sensitivity}  
\end{figure}

In addition to Equation~\eqref{equ:r1} and ~\eqref{equ:r3},  the detailed exploration is obtained with a more specific control on its range and attenuation rate to prioritize early-stage exploration while gradually shifting focus to exploitation, enabling stable policy convergence:
\begin{equation}\label{equ:r2}
    \left\{
    \begin{aligned}
    &r(q, o) = \|f_{\theta_P}(q, o) - f_{\theta_T}(q, o)\|_2^2\\
    &R^\star_{1}(q, o) = \alpha \cdot \frac{r(q, o)-\min\limits_{(q, o)\in B} (r(q, o))}{\max\limits_{(q, o)\in B} (r(q, o))-\min\limits_{(q, o)\in B} (r(q, o))}\\
    &R^\star_{2}(q, o) = I_{a\neq o} \cdot R^\star_1(q, o)\\
    &R^\star(q, o) = \frac{\gamma}{\gamma+n} \cdot R^\star_{2}(q, o)
    \end{aligned}
    \right.
    \end{equation}
This process involves two hyperparameters: $\alpha$, which scales the maximum exploration reward intensity, and $\gamma$, which regulates its decay rate. In this section, we visualize the parameter sensitivity experiments on Countdown-4 for \name(GRPO) in Figure~\ref{fig:sensitivity} to analyze the influence of both on the performance of \name.

Our parameter sensitivity experiments and experiments on other datasets reveal three key insights: (1) Optimal $\alpha$-$\gamma$ combinations vary across datasets due to factors like task complexity. In this paper, $\alpha=0.5$ and $\gamma=40$ demonstrate robust performance as averaged defaults across the three datasets. (2) Extreme $\alpha$ values degrade performance--insufficient $\alpha$ limits exploration, while excessive $\alpha$ (especially with value $\gg1$) risks overemphasizing exploration over correctness (e.g., exploration rewards surpassing the maximum value of outcome-based rewards in some samples). Notably, \name with $\alpha=1.3$ and $\gamma=100$ still outperforms vanilla GRPO ($0.3872$), suggesting tolerance to moderate exploration emphasis. (3) Both slow ($\gamma \gg40$) and rapid ($\gamma \ll40$) decay rates harm performance: slow decay impedes training convergence, while rapid decay prematurely terminates exploration. This effect amplifies with larger $\alpha$ values. 

These findings underscore \name's stability within practical parameter ranges while highlighting the necessity of balanced exploration-exploitation dynamics.

\section{Group Size Analysis}\label{sec:group}
Here we observe the influence of different group size in \name(GRPO). Specifically, we test \name(GRPO) with different group size on Countdown-4. From the result in Table~\ref{tab:group} we could conclude that, as the group size increases, the performance of \name(GRPO) steadily improves over GRPO, which proves the compatible effectiveness of group sampling and \name.

\begin{table}[h]
\centering
\caption{Group size analysis.}
\vspace{-2mm}
\centering
\scalebox{0.85}{
\begin{tabular}{ccc}
    \toprule
    Group Size & GRPO  & \name(GRPO) \\
    \midrule
    2 &   0.1746    &  0.1909 \\
    3 &   0.3374    &  0.3515 \\
    4 &   0.3593    &  0.4261 \\  
    5 &    0.3872   & 0.4739 \\
    \bottomrule
    \end{tabular}%
    }
\label{tab:group}
\vspace{-2mm}
\end{table}

% together with the checklist!!!
\section{Limitations}\label{sec:limitation}
While \name successfully encourages LLMs to explore novel responses, we believe that incorporating more diverse reward functions to evaluate the multi-faceted value of different responses and enabling multi-angle exploration could further enhance reasoning performance. As a future direction, we aim to better analyze these differences to guide LLMs in exploring large action spaces more effectively and developing more complex reasoning capabilities.

\end{document}